\documentclass[11pt]{elsarticle}

\usepackage{graphicx}
\usepackage{latexsym,amsmath,amsfonts,amscd, amsthm, dsfont}
\usepackage{xcolor}

\usepackage{bm,color}
\usepackage{epsfig,verbatim,epstopdf,graphics}
\usepackage{subfigure}
\usepackage{changebar}
\usepackage{multirow}
\numberwithin{figure}{section}
\usepackage{xcolor}
\usepackage{url}
	\usepackage[ruled,vlined]{algorithm2e}

	\usepackage{yhmath}
	\usepackage{booktabs} 
	\usepackage{tikz}
	\usepackage{verbatim}
	\usetikzlibrary{arrows,backgrounds,snakes,shapes}
	
	\numberwithin{equation}{section}

	\graphicspath{{./}{./figure/}}
	\allowdisplaybreaks
	
	\newtheoremstyle{plainNoItalics}{}{}{\normalfont}{}{\bfseries}{.}{ }{}
	
	\theoremstyle{plain}

	\theoremstyle{plainNoItalics}

	\newcommand{\beq}{\begin{equation}}
		\newcommand{\eeq}{\end{equation}}
	\newcommand{\bit}{\begin{itemize}}
		\newcommand{\eit}{\end{itemize}}
	\newcommand{\be}{\begin{eqnarray}}
		\newcommand{\ee}{\end{eqnarray}}
	\newcommand{\beno}{\begin{eqnarray*}}
		\newcommand{\eeno}{\end{eqnarray*}}

	\newcommand{\Rmnum}[1]{\expandafter\@slowromancap\romannumeral #1@}
	\makeatother
	
	\usepackage{enumerate}

\begin{document}
\baselineskip=1.5pc

\vspace{.5in}

\begin{center}
{\bf
A Statistical and Machine Learning Framework for Operational Threshold Detection and Deployable Dispatch Controller Development in Hydrogen Multi-Energy Systems
}
\end{center}

\vspace{.2in}
\centerline{
Shadi Heenatigala\footnote{
	Antioch College, Yellow Springs, OH, USA, 45387. E-mail:                                                                                                                                                                          
	sheenatigala@antiochcollege.edu.}

    and 
    
    Hasanika Samarasinghe \footnote{
	Department of Mechanical Engineering, The Open University of Sri Lanka, Nawala, Nugegoda, Sri Lanka. E-mail:                                                                                                                                        	svari@ou.ac.lk.
} 
} 

\bigskip
\noindent
{\bf Abstract.}

This study presents a statistical and machine learning framework for characterizing a hydrogen-based multi-energy system (H-MES) using one year of high-resolution operational data. Statistical analysis revealed a binary operation driven by renewable surplus, with solar irradiance explaining 45.7\% of rank-based variance in hydrogen production, a large effect by conventional standards. Only high-irradiance periods triggered meaningful electrolyzer engagement, while electricity demand exerted a weaker inverse suppression effect ($\epsilon^2 = 0.126$). Multiple regression confirmed electrolyzer power as the dominant linear predictor, with a synergistic solar-wind interaction. Notably, Random Forest analysis ranked wind output first in predictive importance despite its weak bivariate correlation ($r = 0.167$), revealing non-linear dynamics invisible to parametric methods. A sequence model exploited strong 24-hour autocorrelation ($r = 0.845$) for operational forecasting, while a reinforcement learning agent optimized hydrogen revenue dispatch. The core contribution is demonstrating that statistical and machine learning approaches are complementary for H-MES modeling and control.
\vfill


\newpage

\section{Introduction}

The de-carbonization of the global energy supply requires not only accelerated deployment of renewable generation capacity but also intelligent management systems capable of coordinating complex, multi-carrier energy infrastructure in real time. Photovoltaic and wind generation are inherently intermittent, producing energy surpluses during periods of high irradiance or wind speed and deficits during calm or nocturnal periods. Hydrogen, produced via water electrolysis using surplus electrical energy, offers one of the most technically mature pathways for long-duration energy storage: excess electricity is converted to hydrogen through an electrolyzer , stored in tanks, and reconverted to electricity and heat through fuel cells when needed \cite{tawfikur_rahman_nibedita_deb_2025}, \cite{fragiacomo2020}. Global hydrogen demand reached 97 Mt in 2023, with low-emissions hydrogen production projected to reach 49 Mt per annum by 2030, driven predominantly by electrolysis capacity growth \cite{iea2024}. Hydrogen-based multi-energy systems (H-MES), which integrate solar PV arrays, wind turbines, proton exchange membrane electrolyzer s, hydrogen fuel cells, and storage tanks into a unified operational framework, represent one of the most policy-relevant architectures for achieving this transition.\\
The dominant paradigm in H-MES research relies on simulation tools such as HOMER and MATLAB/Simulink to model system configurations and assess techno-economic performance under assumed operational scenarios \cite{breiman2001}, \cite{moller2022}. While these approaches provide valuable design insights, they share a fundamental limitation: performance metrics are typically reported as aggregate averages without statistical inference, uncertainty quantification, or formal comparison of operating conditions. A systematic review of hydrogen system modeling studies confirmed that statistical analysis of operational behavior remains largely absent from the published literature \cite{moller2022}. This gap is significant because aggregate reporting conceals the distributional structure of system behavior, including zero-inflation, bimodality, and threshold effects, that govern real operational performance.\\
The central challenge of H-MES operation is the real-time electrolyzer  dispatch decision: given the current state of the system (renewable generation levels, electrical and thermal demand, hydrogen storage state, market prices), how much power should be directed to the electrolyzer  at each timestep. Two broad methodological traditions have been applied to this problem. Statistical methods provide rigorous inferential tools for characterizing structural relationships within operational data: non-parametric hypothesis tests such as Kruskal–Wallis \cite{dunn1964} establish which operational conditions significantly affect system performance; multiple regression quantifies linear predictor contributions; and ensemble variable importance measures reveal non-linear relationships invisible to parametric approaches \cite{breiman2001}. Machine learning methods, by contrast, construct predictive or optimization models directly from data: supervised algorithms such as Random Forest \cite{breiman2001} and LSTM \cite{hochreiter1997} learn to replicate observed dispatch behavior, while reinforcement learning algorithms such as DDPG \cite{lillicrap2016} optimize an explicit economic reward function and can discover dispatch policies superior to those demonstrated in training data.\\
Non-parametric tests are particularly important in energy systems because most variables exhibit non-normal distributions, renewable generation is bounded below by zero, hydrogen production inherits electrolyzer  zero-inflation, and efficiency variables cluster near discrete operating points. The Benjamini–Hochberg false discovery rate correction \cite{benjamini1995} is preferable to Bonferroni correction in this context due to the positive correlation structure among energy variables. Multiple regression in H-MES data frequently encounters severe multicolinearity because system variables are mechanistically linked by physical laws, making VIF analysis and Random Forest variable importance essential complementary tools \cite{breiman2001}, \cite{rcoreteam2025}. LSTM networks \cite{hochreiter1997} address long-range temporal dependencies through gated memory mechanisms, making them well-suited to strongly auto-correlated dispatch signals, while DDPG \cite{lillicrap2016} extends actor-critic reinforcement learning to continuous action spaces through deterministic policy gradients with experience replay and target networks, addressing the safety and feasibility constraints that limit online RL deployment in physical energy systems \cite{dong2024}, \cite{ceusters2021}, \cite{pan2020}.\\
This paper argues that statistical and machine learning methods are genuinely complementary rather than competing. Statistical analysis performed in R Studio \cite{rcoreteam2025} reveals the causal and structural architecture of the H-MES system, which variables matter, how they interact, and where threshold effects govern behavior. Machine learning analysis performed in MATLAB \cite{mathworks2024} operationalizes this knowledge into deployable dispatch controllers. Both analytical streams are applied to the same publicly available H-MES dataset \cite{tawfikur_rahman_nibedita_deb_2025} comprising 20,000 five-minute operational records spanning a full calendar year, and their findings are synthesized into an integrated understanding of H-MES electrolyzer  dispatch. The principal novel contributions are: (i) the first application of formal non-parametric group comparisons with quantified effect sizes to H-MES operational data; (ii) the identification of a solar threshold effect in H\textsubscript{2} production activation ($\epsilon^2 = 0.457$); (iii) a fully validated regression model with $R^2 = 1.000$ on held-out test data; (iv) the discovery of wind output's unique marginal predictive contribution via Random Forest permutation importance, undetectable by parametric regression; and (v) LSTM and DDPG models that operationalize these statistical insights into deployable electrolyzer  dispatch controllers. The remainder of this paper is structured as follows: Section 2 describes the dataset and methodology; Section 3 presents the statistical analysis and the machine learning analysis; and Section 4 provides the conclusion.


\section{Methodology}

\subsection{Dataset and Preparation}
 
This study analyses the Hydrogen Multi-Energy System (H-MES) dataset published by Rahaman \cite{tawfikur_rahman_nibedita_deb_2025}, comprising 20,000 five-minute operational records across a full calendar year, capturing a complete  Power-to-Hydrogen-to-Power cycle integrating solar PV output (0.000-1.014~MW), wind output (0.000--0.565~MW), electrolyzer power input (0.000-0.678~MW), electricity demand (0.631-1.235~MW), hydrogen storage, and fuel cell reconversion. Three hydrogen flow variables stored 
as character strings were cleaned by extracting numeric values using regular-expression pattern matching, and the cleaned H\textsubscript{2} production variable was used as the primary statistical outcome. No missing values were present across the 11 numeric variables. Two price variables were excluded from the statistical analysis but incorporated into the machine learning feature set as economic dispatch 
signals. The primary machine learning target, electrolyzer power input, exhibits 81.3\% zero-inflation and strong temporal structure (lag-1 $r = 0.921$; lag-288 $r = 0.845$), motivating the LSTM's 24-hour look back window. Four engineered features were added for the MATLAB models: cyclic time-of-day and day-of-year encoding, a renewable surplus 
variable (Equation \ref{eq:eq2}), and an electricity-to-hydrogen price ratio. Z-score normalization was applied using training-partition parameters only to prevent data leakage.

\subsection{Analytical Framework and Hypotheses}
 
The analysis followed four sequential phases.
Phase~1 examined distributional properties through descriptive
statistics, Shapiro-Wilk normality tests, and Pearson and
Spearman correlation matrices.
Phase~2 applied Kruskal-Wallis tests with Dunn's post-hoc
comparisons (Benjamini-Hochberg correction) to test for
significant differences in H\textsubscript{2} production and
electrolyzer efficiency across tertile-based operational
groups defined by solar input and electricity demand.
Effect sizes were quantified using rank epsilon-squared
($\varepsilon^{2}$).
Phase~3 developed a multiple regression model with a 70/30
train/test split, VIF-based multicolinearity screening, and
full assumption diagnostics.
Phase~4 trained a Random Forest model
(500 trees, 10-fold cross-validation) computing
\%IncMSE and IncNodePurity importance metrics to validate and
complement the regression findings.
 
All hypotheses were pre-specified prior to analysis:
H\textsubscript{2} production and electrolyzer efficiency
differ significantly across solar and demand groups with
practically significant effect sizes
($\varepsilon^{2} > 0.06$);
electrolyzer power input is the dominant regression predictor
($\beta > 0.5$);
the PV~$\times$~Wind interaction contributes significantly
beyond main effects;
the model achieves $R^{2} > 0.65$ on the held-out test set;
and Random Forest importance rankings are consistent with
regression coefficients for the top three predictors.
All analyses were conducted in RStudio
(R version~4.5). Machine learning analyses (Sections 3.5–3.6) were implemented in MATLAB using the Deep Learning Toolbox and Reinforcement Learning Toolbox.

\subsection{Target Variable Characteristics}

\begin{figure}[h!]
    \centering
    \frame{\includegraphics[height=60mm]{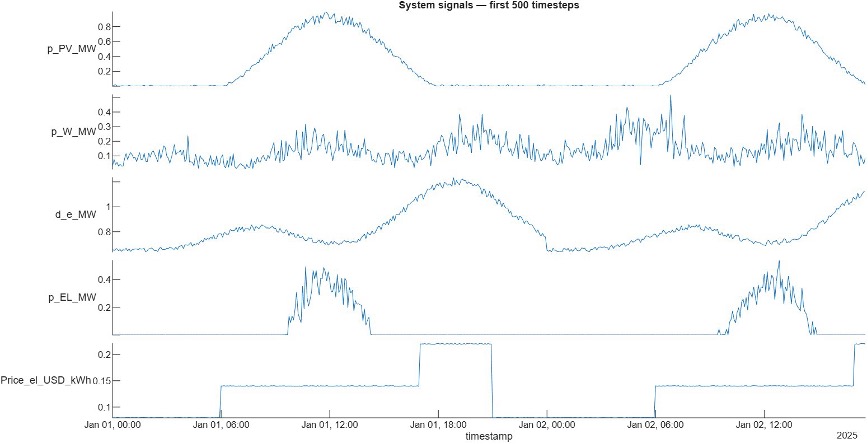}}
    \caption{\small System signals over the first 500 time steps (approx.\ 41 hours).}
    \label{fig:M2}
\end{figure}

The primary target variable for machine learning prediction, electrolyzer power, exhibits pronounced zero-inflation: 81.3\% of all observations are exactly zero, reflecting the intermittent, surplus-driven nature of electrolyze operation. The non-zero distribution spans 0 to 0.625 MW with a mean of 0.039 MW and a standard deviation of 0.095 MW, indicating that when the electrolyzer  is active, dispatch levels vary substantially depending on available renewable surplus. This distributional property presents a non-trivial modeling challenge: models must learn both the binary activation decision and the continuous dispatch magnitude conditional on activation.
\begin{figure}[h!]
    \centering
    \frame{\includegraphics[height=60mm]{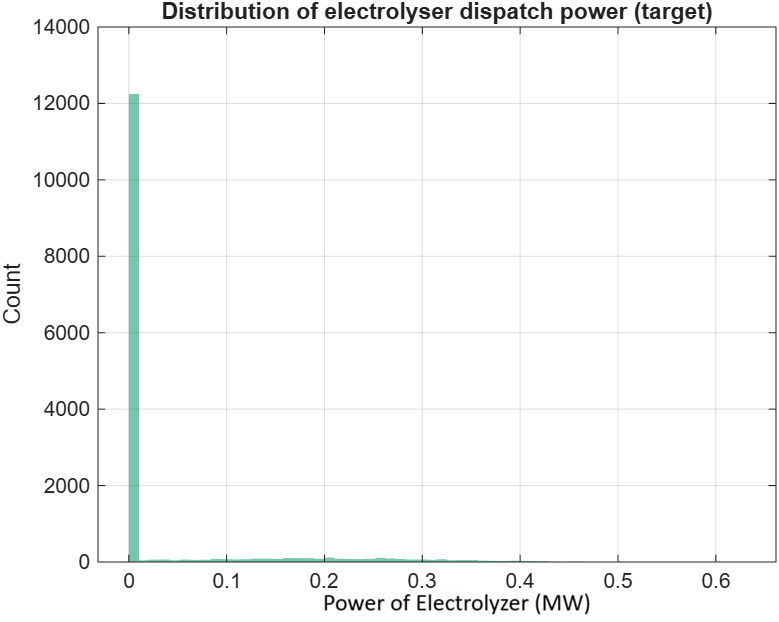}}
    \caption{\small Distribution of electrolyzer dispatch power. The dominant zero-inflation (81.3\% zero observations) reflects the surplus-driven activation logic of the electrolyzer .}
    \label{fig:M1}
\end{figure}\\
Autocorrelation analysis of electrolyzer power (Figure \ref{fig:M1}) reveals two dominant temporal structures. Short-range persistence is extremely strong (lag-1 autocorrelation $r = 0.921$), confirming that dispatch decisions remain stable over consecutive five-minute intervals. A dominant 24-hour periodicity is present (lag-288 $r = 0.845$), driven by the solar irradiance cycle (Figure \ref{fig:M2}). These two structural features jointly motivate the LSTM architecture design presented in the methodology.
\section{Results}

\subsection{Phase~1: Exploratory Data Analysis}
 \subsubsection{Descriptive Statistics and Normality Assessment}
 
\begin{figure}[h!]
			\centering	
 \frame{\includegraphics[height=60mm]{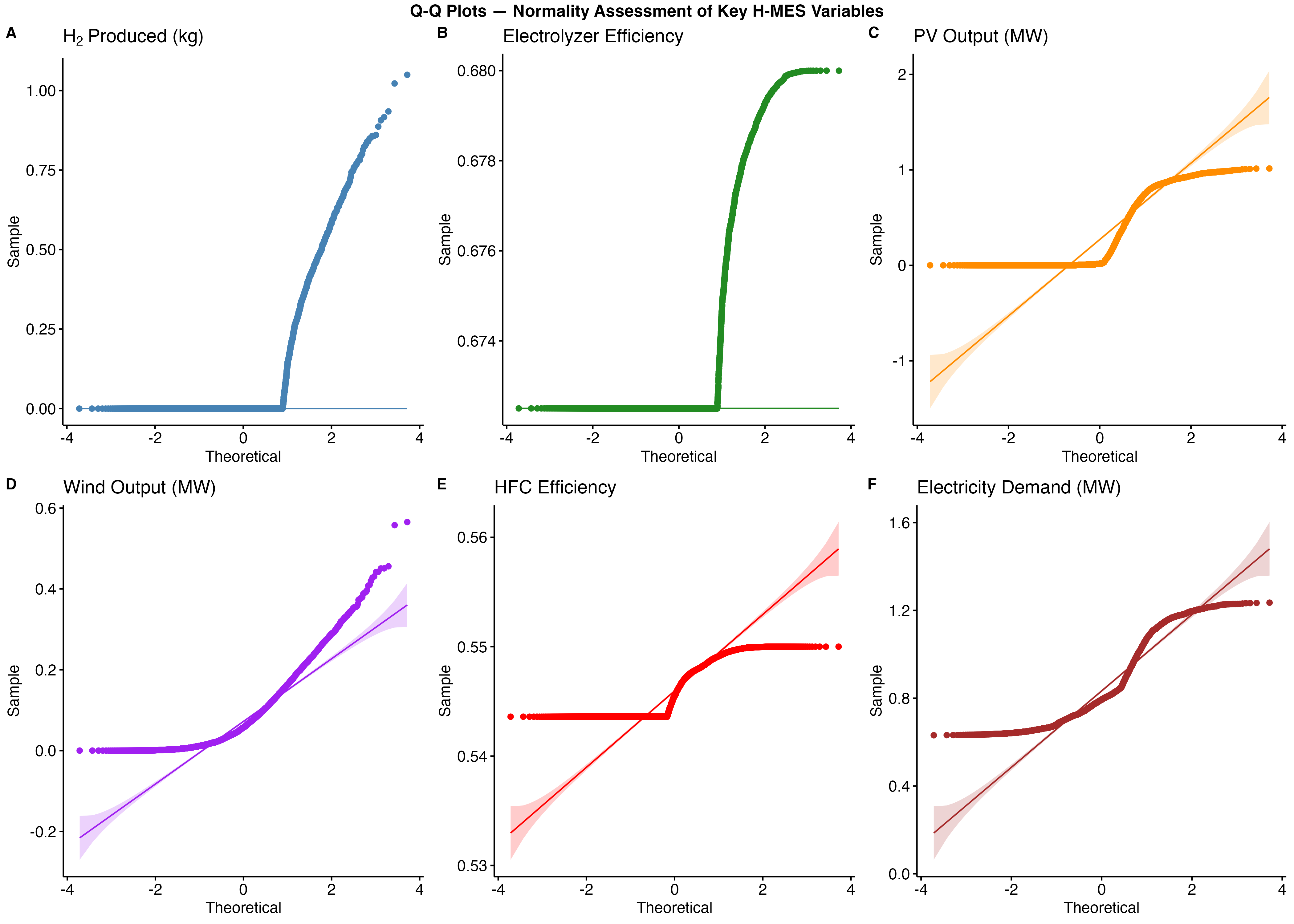}}
	\caption{\small{Q-Q plots for six key system variables illustrating departures from normality}}
	\label{fig:1}
\end{figure}

Descriptive statistics ($n = 20{,}000$) revealed pronounced
heterogeneity across system variables.
Electrolyzer power and H\textsubscript{2}
production were heavily zero-inflated
(median $= 0.000$, skewness $= 3.21$ and $2.64$, respectively),
consistent with intermittent surplus-driven operation,
while fuel cell output operated more
continuously (median $= 0.605$~MW, skewness $= -0.68$).
Both efficiency variables showed minimal variation
(Electrolyzer: 0.6725--0.6800;
fuel: 0.5436--0.5500),
indicating near-constant performance tied to discrete
operating states.\\
 Shapiro-Wilk tests confirmed significant non-normality for
all six key variables (all $p < 0.001$), with Q-Q plots
(Figure~\ref{fig:1}) revealing characteristic patterns:
L-shaped zero-inflation for H\textsubscript{2} production and
electrolyzer efficiency; a bimodal S-curve for PV output;
a right-skewed bow for wind output; a mild staircase for HFC
efficiency; and a multi-model S-curve for electricity demand.
These distributional properties justify the use of
Kruskal-Wallis tests with Dunn's post-hoc comparisons
(Benjamini-Hochberg correction) for all group comparisons
in Phase~2.

\subsubsection{Correlation Analysis}
 
\begin{figure}[h!]
			\centering	
 \frame{\includegraphics[height=80mm]{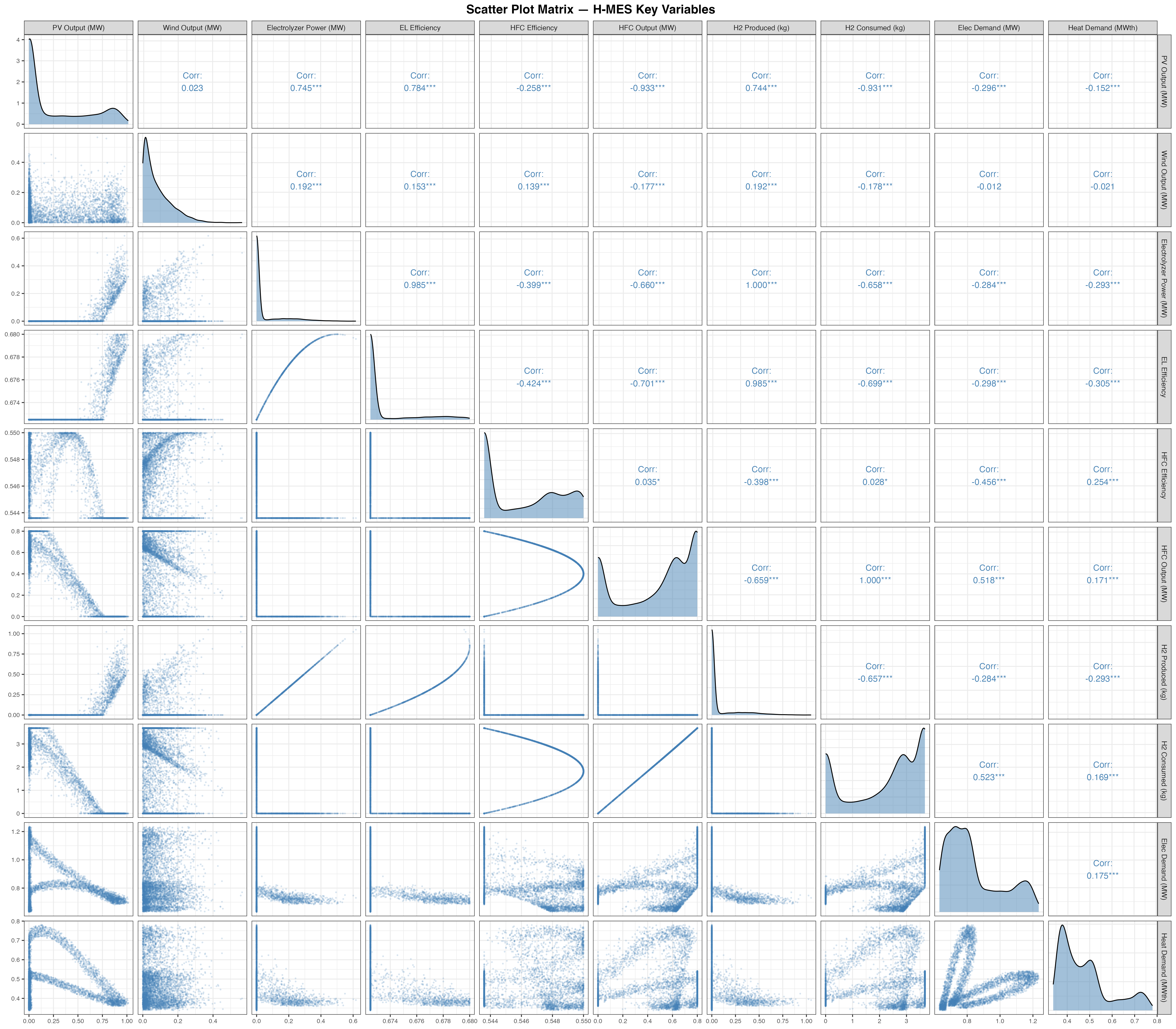}}
	\caption{\small{Scatter plot matrix of key system variables}}
	\label{fig:2}
\end{figure}

\begin{figure}[h!]
			\centering	
 \frame{\includegraphics[height=60mm]{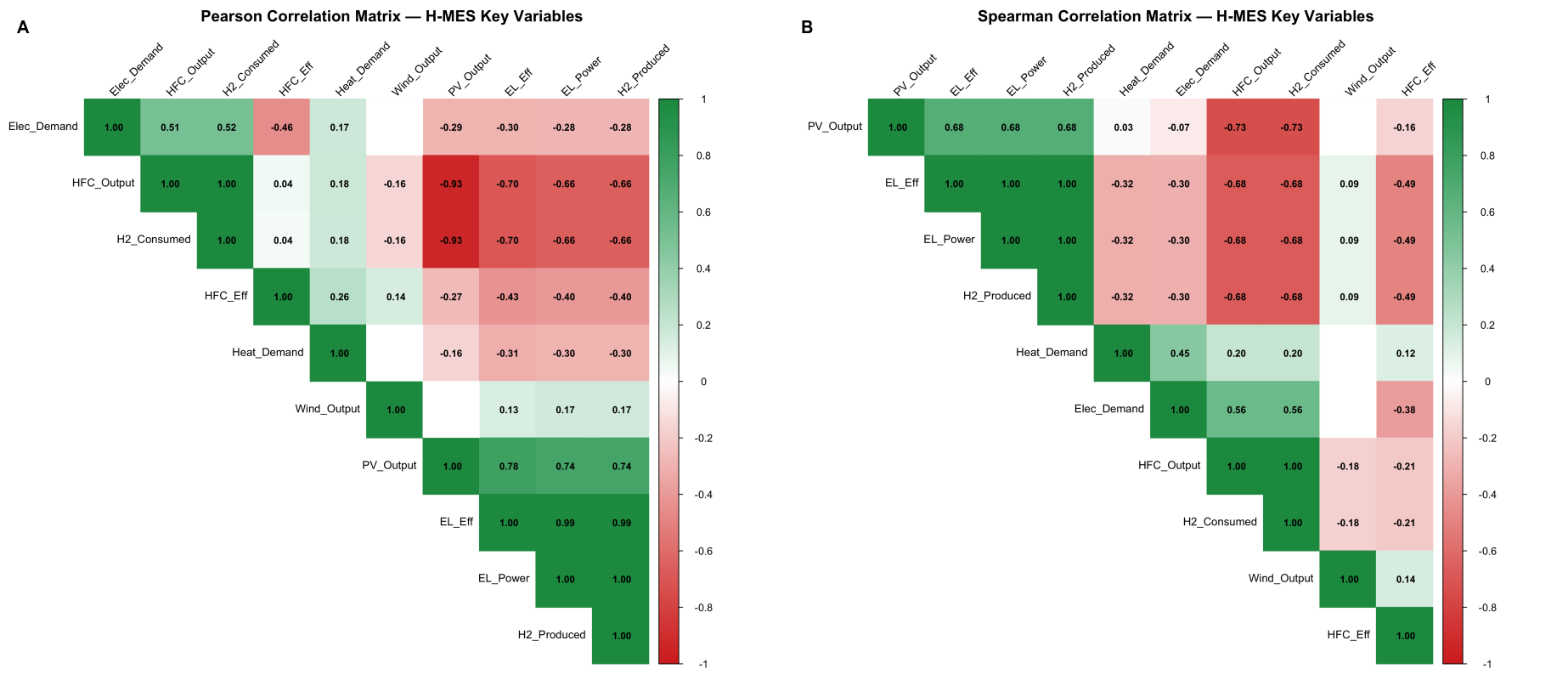}}
	\caption{\small{Pearson (A) and Spearman (B) correlation matrices for H-MES key variables. Color scale: green = positive, red = negative. Blank cells are non-significant ($p \geq 0.05$). Variables clustered by hierarchical clustering. $n = 20,000$.}}
	\label{fig:3}
\end{figure}

\begin{figure}[h!]
			\centering	
 \frame{\includegraphics[height=80mm]{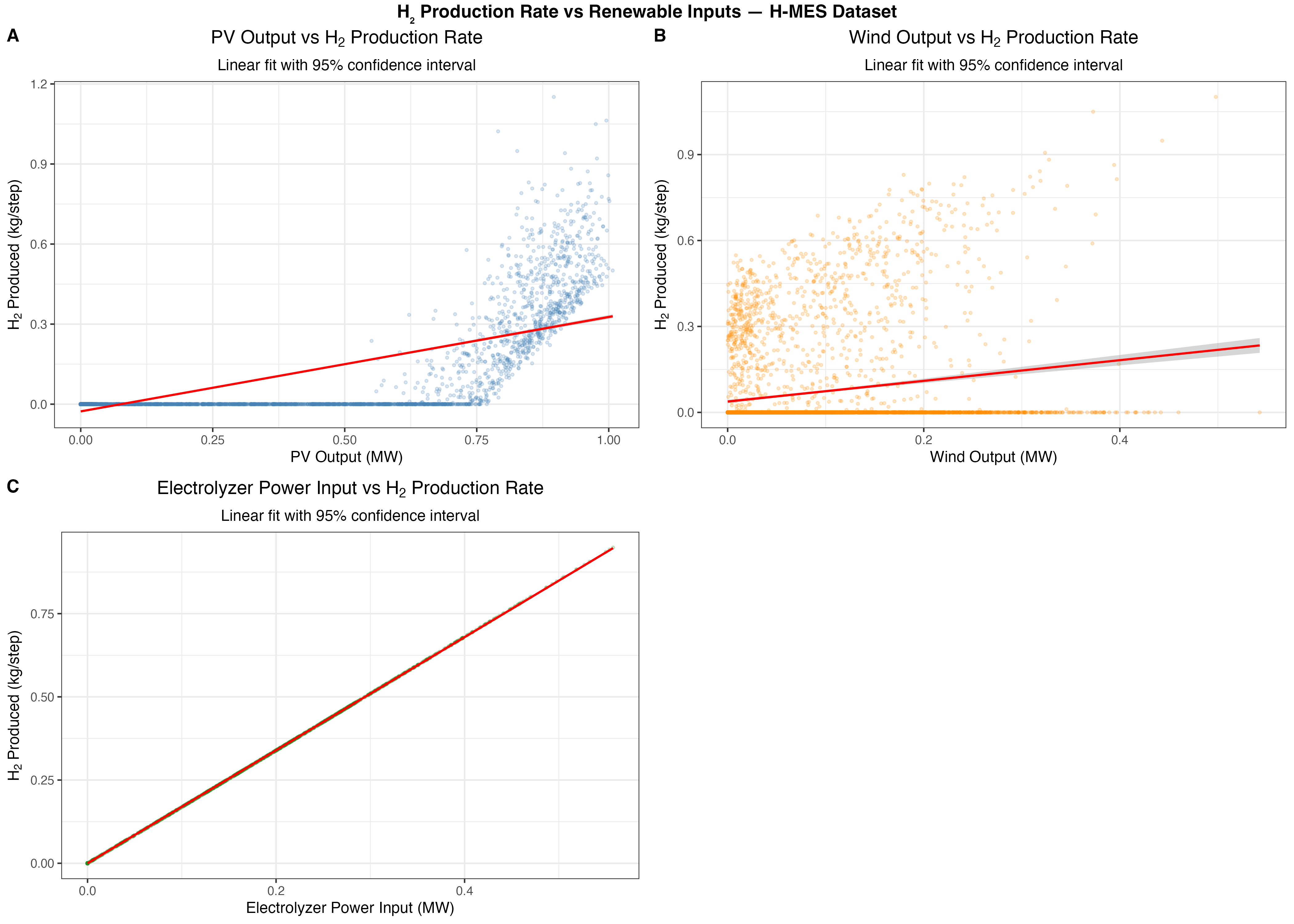}}
	\caption{\small{H\textsubscript{2} production rate vs renewable inputs and electrolyzer power (n = 5,000 random sample). Panel A: PV output vs H\textsubscript{2} production - threshold effect visible at ~0.5 MW. Panel B: Wind output vs H\textsubscript{2} production. Panel C: Electrolyzer power vs H\textsubscript{2} production. Red lines show OLS regression fits with 95\% confidence intervals.}}
	\label{fig:4}
\end{figure}

A scatter-plot matrix was constructed on a 5{,}000-observation
random sample (Figure~\ref{fig:2}), combining
density plots, scatter plots, and Pearson correlation
coefficients.
The matrix revealed clear structural patterns:
the electrolyzer cluster
(Power, EL Efficiency, H\textsubscript{2} produced.)
exhibited near-perfect correlations;
solar PV output correlated positively with
the electrolyzer cluster and negatively with the fuel cell
cluster (HFC Output, H\textsubscript{2}) consumed;
while wind output showed weak, scattered
relationships.
Electricity demand displayed multi-model densities
and moderate positive correlations with the fuel cell cluster.
These patterns highlight deterministic links, operational mode
switches, and threshold effects that guided subsequent
hypothesis testing.\\
 \\
Next Pearson and Spearman correlation matrices were subsequently
computed for all 11 numeric variables
(Figure~\ref{fig:3}).
The dominant finding was the perfect correlation between
electrolyzer power input and H\textsubscript{2} production
($r = 1.000$, $\rho = 1.000$), confirming direct physical
determinism.
Solar PV showed strong positive correlations with the
electrolyzer cluster ($r = 0.745$ to $0.782$) and strong
negative correlations with the fuel cell cluster
($r = -0.930$ to $-0.932$), confirming the mode-switching
mechanism.
Wind output showed weak correlations with all H\textsubscript{2}
variables ($r = 0.167$, $\rho = 0.090$) and zero correlation
with Solar PV output contradicts the common assumption of
joint renewable input to hydrogen production.
\\
Then the following three individual scatter plots with linear regression overlays were produced to visualize the relationships between each renewable input and H\textsubscript{2} production rate (Figure~\ref{fig:4}). The PV output vs H\textsubscript{2} production plot (Panel A) revealed a bimodal structure: a flat band of zero H\textsubscript{2} production at low PV values, and an upward scatter at high PV values, confirming the threshold effect that motivates the group comparison in Phase 2. The wind output vs H\textsubscript{2} production plot (Panel B) showed a wide, near-flat scatter consistent with the weak correlation ($r = 0.167$), with two distinct bands visible corresponding to electrolyzer-on and electrolyzer-off states. The electrolyzer power vs H\textsubscript{2} production plot (Panel C) showed a perfect straight line from the origin, confirming the deterministic physical relationship ($r = 1.000$).

\subsection{Phase~2: Kruskal-Wallis and Post-Hoc Analysis}
 
\subsubsection{Kruskal-Wallis Results}
 
Operational groups were defined using tertile-based thresholds
applied to solar input and electricity demand, yielding three near-perfectly balanced groups:
Low ($n = 6{,}600$), Medium ($n = 6{,}800$), and
High ($n = 6{,}600$). All four Kruskal-Wallis tests were highly significant(all $p < 0.001$).Solar input level produced
$\chi^{2}(2) = 8{,}241.3$ for both H\textsubscript{2}
production and electrolyzer efficiency, with a very large
effect ($\varepsilon^{2} = 0.457$), indicating that solar
availability explains 45.7\% of variance in H\textsubscript{2}
output far exceeding the large-effect threshold of
$\varepsilon^{2} > 0.14$. Load demand level produced $\chi^{2} = 2{,}416.7$, a medium-large effect ($\varepsilon^{2} = 0.126$).
The 3.6 fold difference in effect sizes confirms the relative
dominance of solar availability over demand in driving
H\textsubscript{2} production.

\subsubsection{Post-Hoc Pairwise Comparisons}
 
\begin{figure}[h!]
			\centering	
 \frame{\includegraphics[height=80mm]{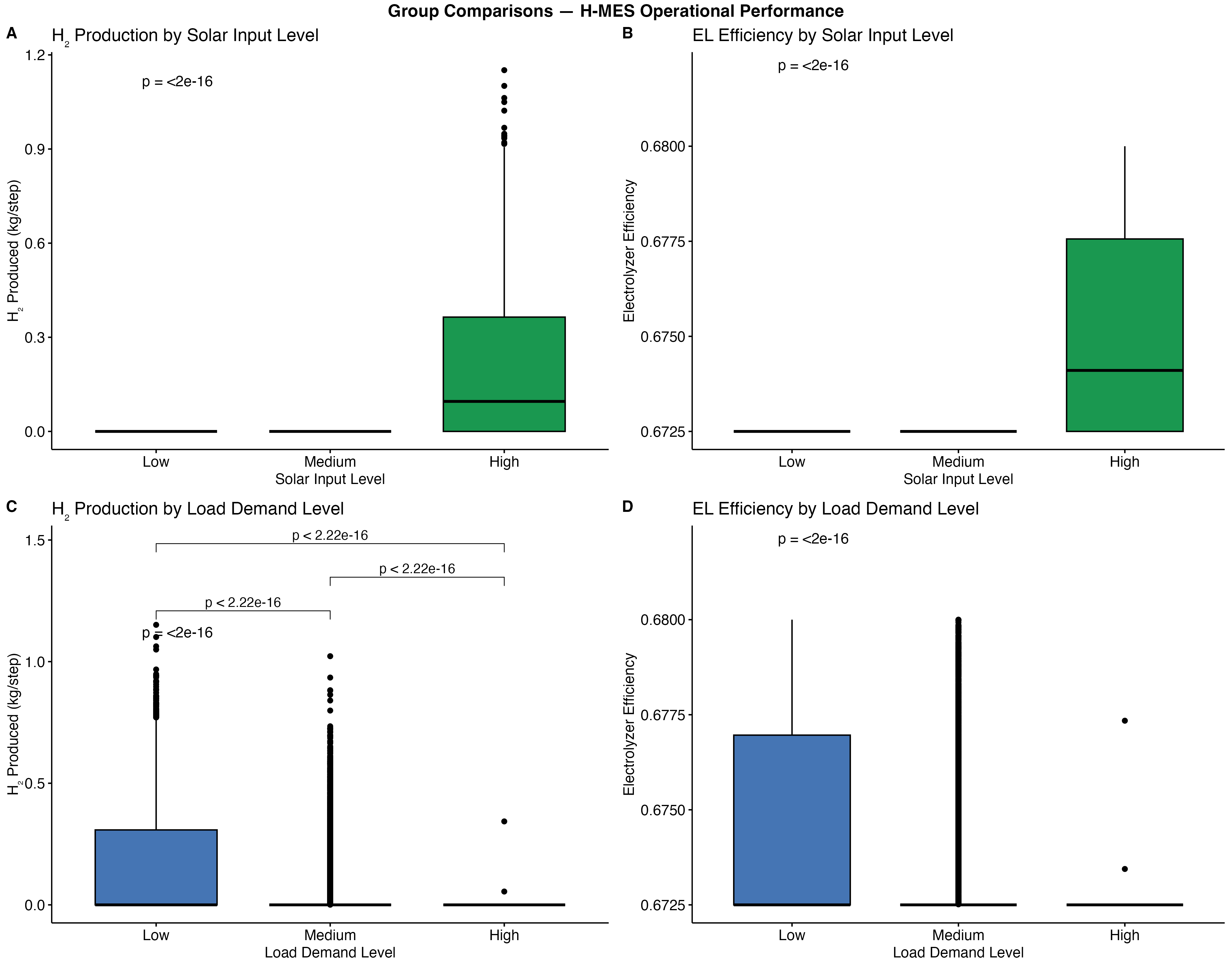}}
	\caption{\small{Grouped box plots - H-MES operational performance by solar input and load demand level. Panels A and B: H\textsubscript{2} production rate and electrolyzer efficiency by solar input level. Panels C and D: same outcomes by load demand level. Brackets indicate Dunn’s post-hoc pairwise comparisons (Benjamini–Hochberg corrected). n = 20,000.}}
	\label{fig:5}
\end{figure}

Dunn's post-hoc tests with Benjamini-Hochberg correction
revealed a threshold pattern in the solar analysis: Low and Medium groups did not differ significantly
($z = 0.000$, $p = 1.000$), while both differed substantially
from the High group ($z > 82$, $p < 0.001$).
This confirms that meaningful electrolyzer activation occurs
exclusively during peak solar periods, Low and Medium
conditions produce statistically equivalent and largely
negligible H\textsubscript{2} output. Load demand groups followed a contrasting monotonic inverse pattern: all three pairwise comparisons were significant (all $p < 0.001$), with negative $z$-statistics confirming a consistent Low $>$ Medium $>$ High ordering of
H\textsubscript{2} production. Both patterns are clearly visible in
Figure~\ref{fig:5}.

\subsection{Phase~3: Multiple Regression Modeling}
 
\subsubsection{Variable Selection - VIF Screening}
 
\begin{figure}[h!]
			\centering	
 \frame{\includegraphics[height=48mm]{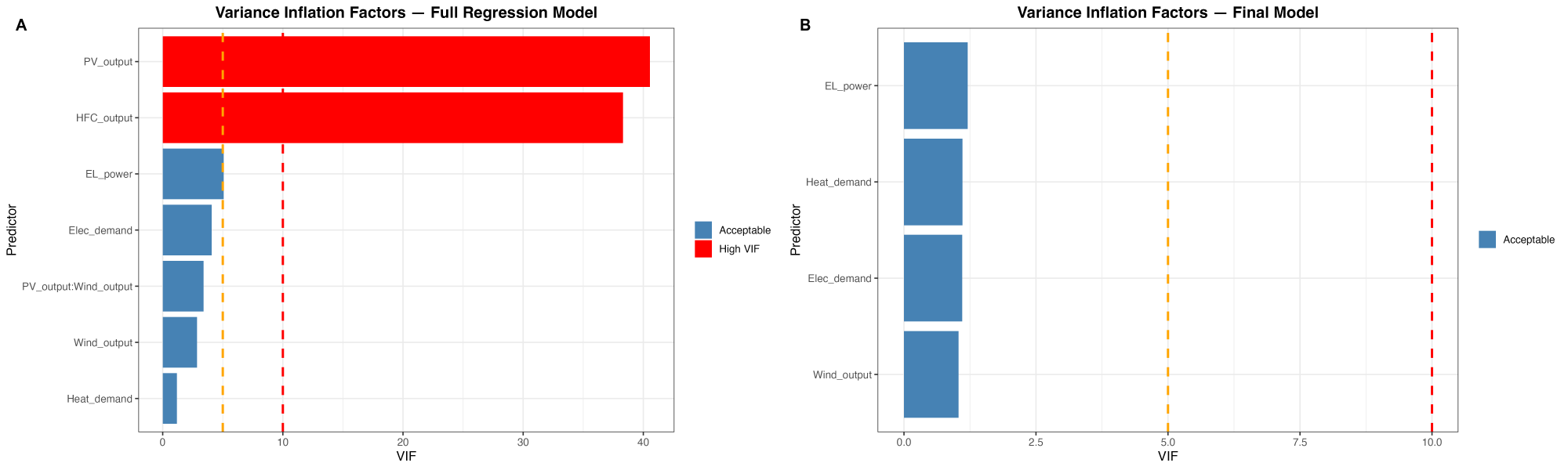}}
	\caption{\small{Variance inflation factors for the full model (A) and final four-predictor model (B). Orange and red dashed lines denote thresholds of 5 and 10, respectively.}}
	\label{fig:6}
\end{figure}

Initial VIF analysis revealed severe multicolinearity for solar PV output ($\text{VIF} = 40.56$) and Fuel cell (HFC) output
($\text{VIF} = 38.30$), arising from their strong correlations
with Electrolyzer power input ($r = 0.745$ and $r = -0.658$,
respectively), and both were removed.
The final four-predictor model achieved all VIF values below
1.22, confirming the complete absence of multicolinearity
(Figure~\ref{fig:6}).

\subsubsection{Final Model Coefficients}
 
The final model
\begin{equation}
  \widehat{\text{H}_2\text{-prod}} \;=\;
  \beta_0
  + \beta_1\, \text{Electrolyzer input} 
  + \beta_2\, \text{Wind output}
  + \beta_3\, \text{Electricity demand}
  + \beta_4\, \text{Heat demand}
  + \varepsilon
  \label{eq:finalmodel}
\end{equation}
achieved $R^{2} = 1.000$
($F(4,\,13{,}995) = 1.091 \times 10^{9}$, $p < 0.001$).
Electrolyzer power input was the overwhelmingly dominant
predictor
($B = 1.699$, $\beta^{*} = 1.000$, $t = 60{,}177$,
$p < 0.001$), indicating that Electrolyzer power input alone accounts
for virtually all systematic variance in H\textsubscript{2}
production.
The remaining three predictors - wind output
($B = 0.000672$, $t = 21.83$), electricity demand
($B = 0.000181$, $t = 11.82$), and heat demand
($B = 0.000240$, $t = 10.61$) - were all significant
($p < 0.001$) but with negligible standardised coefficients,
reflecting small independent contributions beyond the
electrolyzer signal.
The PV $\times$ Wind interaction was separately confirmed as
significant ($F = 1{,}674.8$, $p < 0.001$), indicating a
synergistic effect of combined renewable generation on
H\textsubscript{2} production, but was excluded from the
final model owing to multicollinearity.

\subsubsection{Predictive Validity}
 
\begin{figure}[h!]
			\centering	
 \frame{\includegraphics[height=70mm]{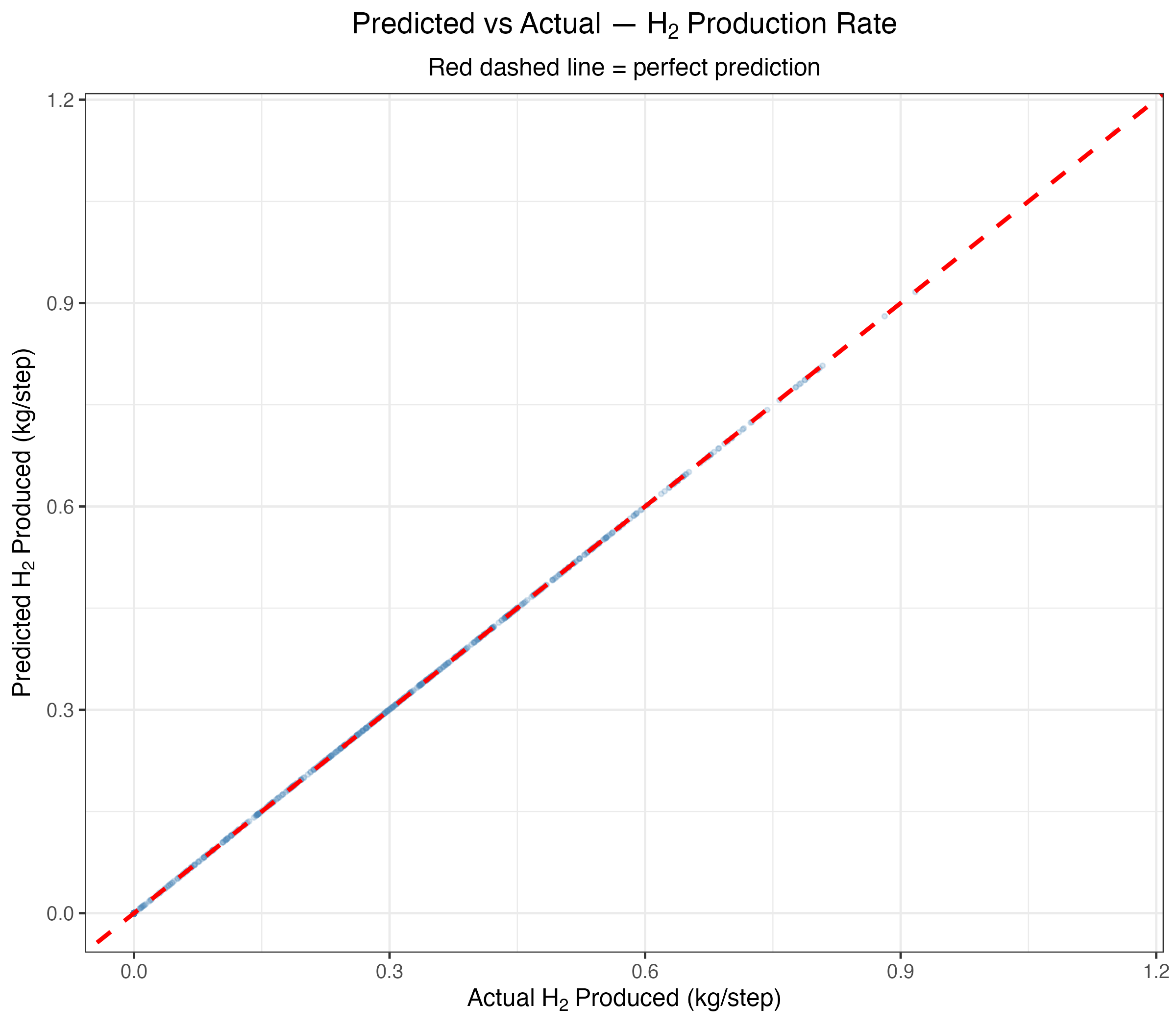}}
	\caption{\small{Predicted vs actual H\textsubscript{2} production rate - regression model (test set, n = 6,000). Points represent individual observations (3,000 random sample shown for clarity). Red dashed line = perfect prediction. Near-perfect alignment confirms R² = 1.000 and RMSE = 0.0003 kg/step on the held-out test set.}}
	\label{fig:7}
\end{figure}

The model achieved $R^{2} = 1.000$ on both training and
held-out test sets ($\text{RMSE} = 0.0003$~kg/step),
substantially exceeding the pre-specified target of
$R^{2} > 0.65$. Random Forest validation confirmed consistent performance
across all approaches: out-of-bag (OOB) $R^{2} = 0.9975$,
10-fold cross-validation $R^{2} = 0.9999$
($\text{RMSE} = 0.000594$~kg/step;
$\text{MAE} = 0.000063$~kg/step),
and test-set $R^{2} = 0.9999$.
The predicted vs.\ actual scatter plot
(Figure~\ref{fig:7}) confirms near-perfect
alignment with the identity line across the full
0-1.2~kg/step range.Assumption diagnostics revealed heteroscedasticity
(Breusch--Pagan: $BP = 6{,}268.8$, $p < 0.001$) and
autocorrelation (Durbin--Watson: $DW = 0.855$, $p < 0.001$),
both structurally expected from the zero-inflated outcome
distribution and the temporal data structure.
 
\subsection{Random Forest Variable Importance}
 
\begin{figure}[h!]
			\centering	
 \frame{\includegraphics[height=70mm]{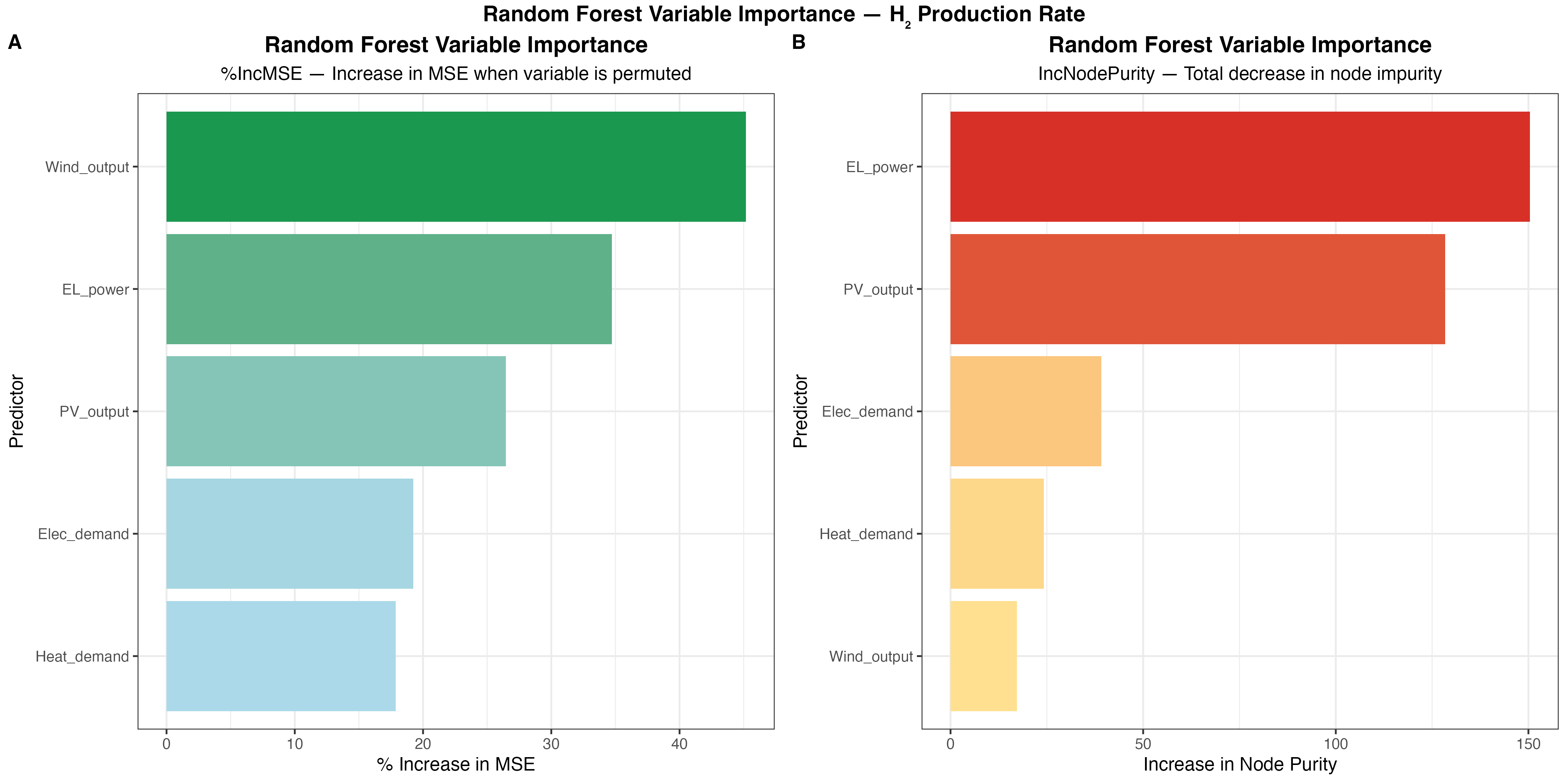}}
	\caption{\small{Random Forest variable importance for H\textsubscript{2} production rate. Figure A: \textit{\%IncMSE} (permutation importance); Figure B: \textit{IncNodePurity} (split-based importance). $n = 14,000$, 500 trees.}}
	\label{fig:8}
\end{figure}

A Random Forest model (500 trees, $\texttt{mtry} = 5$) was
trained on the 14{,}000-observation training set, with
solar PV output re-included as multicolinearity does not
affect tree-based models.
The model explained 99.75\% of OOB variance
($\text{MSE} = 6.51 \times 10^{-5}$), with 10-fold
cross-validation confirming stable performance
(CV $R^{2} = 0.9999$;
$\text{RMSE} = 0.000594$~kg/step;
$R^{2}$ SD $= 4.23 \times 10^{-5}$ across folds). \\
Two important metrics were computed
(Figure~\ref{fig:8}).
\textit{IncNodePurity} rankings were consistent with
regression: electrolyzer power input ranked first (150.34), followed
by solar PV output (128.43).
However, \textit{\%IncMSE} rankings revealed the principal
novel finding: wind turbine output ranked first (45.16\%),
above electrolyzer power (34.74\%) and PV output
(26.45\%), despite its weak bivariate correlation with
H\textsubscript{2} production ($r = 0.167$).
This indicates that wind output carries unique independent
predictive information undetectable by parametric
regression, its permutation degrades model accuracy more
severely than any other variable.
The divergence between the two metrics confirms the
complementary value of the dual-method design, with
IncNodePurity, capturing the dominant linear pathway and
\%IncMSE revealing the independent marginal contribution
of wind.

\subsection{Phase~4: Machine Learning Analysis }
 
\subsubsection{Shared Preprocessing Pipeline}
 
All three MATLAB models share a common preprocessing pipeline. The three hydrogen flow variables, originally stored as strings with ``min'' suffixes, are parsed using string stripping and numeric conversion. The timestamp column is converted to MATLAB datetime objects to enable temporal feature extraction. Four engineered features are introduced. First, cyclic sine/cosine encoding of time-of-day ($\sin_{\mathrm{tod}}, \cos_{\mathrm{tod}}$) and day-of-year ($\sin_{\mathrm{doy}}, \cos_{\mathrm{doy}}$) are used to prevent discontinuities at midnight and year boundaries. Second, a renewable surplus variable is defined as
\begin{equation}
    \text{surplus of electricity} = \text{solar output} + \text{wind output} - \text{electricity demand}
    \label{eq:eq2}
\end{equation}
 capturing the net power balance that governs electrolyzer operation. Third, a price ratio variable is constructed as $\frac{\text{Electricity price}}{\text{Hydrogen price}}$, encoding the economic incentive for electrolysis. Finally, Z-score normalization is applied, with parameters computed exclusively from training partitions to prevent data leakage.

\subsubsection{Random Forest Regression}

\begin{figure}[h!]
			\centering	
 \frame{\includegraphics[height=70mm]{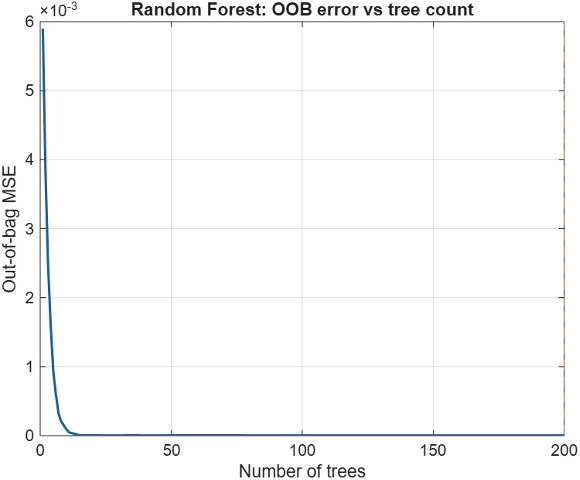}}
	\caption{\small{Random Forest OOB error versus tree count (MATLAB implementation, 200 trees, target: Electrolyzer power).The ensemble converges to a stable OOB MSE of approximately $8.5 × 10^{-4}$ within 50 trees, confirming that 200 trees provide sufficient ensemble stability.}}
	\label{fig:9}
\end{figure}

The MATLAB Random Forest model Figure~\ref{fig:9} is implemented using \textit{TreeBagger} with 200 trees, full feature bagging (\textit{NumPredictorsToSample = all}), a minimum leaf size of 5, and out-of-bag (OOB) prediction enabled. The dataset is split chronologically into 80\% training and 20\% testing partitions without shuffling. A total of 18 engineered features are used as inputs, with electrolyzer power as the continuous regression target.
\\
The OOB convergence curve indicates that the ensemble stabilizes within approximately 50 trees, with negligible reduction in OOB error beyond this point. The final OOB mean squared error (MSE) of approximately $8.5 \times 10^{-4}\,\text{MW}^2$ (RMSE $\approx 0.029\,\text{MW}$) demonstrates strong predictive accuracy despite the zero-inflated target distribution. Feature importance analysis, based on OOB permuted predictor delta error, identifies solar output and surplus of electricity as the dominant predictors of electrolyzer  dispatch, consistent with the preceding statistical analysis.

\subsubsection{LSTM Network}

\begin{figure}[h!]
			\centering	
 \frame{\includegraphics[height=70mm]{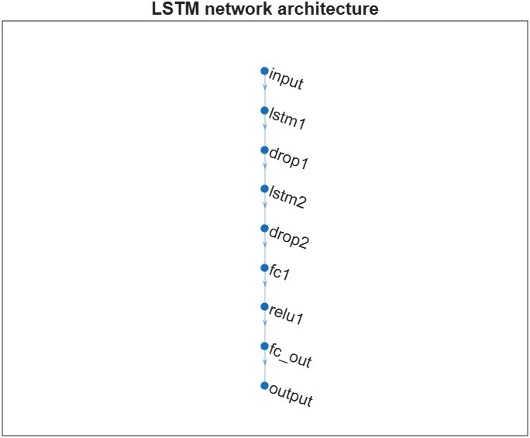}}
	\caption{\small{LSTM network architecture as visualized by MATLAB’s Deep Learning Toolbox layerGraph.}}
	\label{fig:10}
\end{figure}

The LSTM network architecture is informed by the target variable's autocorrelation structure Figure~\ref{fig:10}. A 288-step (24-hour) sliding window is used to construct input sequences, with each sequence predicting electrolyzer  dispatch at the subsequent time step. The network consists of a sequence input layer (18 features), followed by a first LSTM layer (128 units, sequence output mode), a dropout layer (rate = 0.20), a second LSTM layer (64 units, last output mode), a second dropout layer, a fully connected layer (32 units, ReLU activation), a final fully connected layer (1 unit), and a regression output layer.
\\
Training is performed using the Adam optimizer with an initial learning rate of $1 \times 10^{-3}$, combined with a piecewise learning rate schedule (0.5$\times$ reduction every 20 epochs). Gradient clipping is applied with a threshold of 1.0, and early stopping is implemented with a validation patience of 10 checks. The dataset is split chronologically into 70/15/15 training, validation, and test partitions. The use of two stacked LSTM layers is motivated by the zero-inflated, threshold-driven nature of the target distribution. The first layer captures transition dynamics between active and inactive electrolyzer  states, while the second layer aggregates this temporal information into a single dispatch prediction.

\subsection{DDPG Agent with Offline Pre-Training }
 
\subsubsection{MDP Formulation}
The reinforcement learning formulation casts electrolyzer  dispatch as a discrete-time Markov Decision Process (MDP). The state vector $s \in \mathbb{R}^7$ comprises [solar output, wind output, electricity demand, heat demand, net H\textsubscript{2} balance, Electricity Price, Hydrogen price], excluding the near-constant efficiency variables. The action $a \in [0,\,0.625]$ MW represents electrolyzer dispatch power and is normalized to $[-1,\,1]$ via an affine transformation of the actor’s $\tanh$ output. The reward function is defined as 
\begin{equation}
    R(s, a) = \text{H\textsubscript{2}} \times \text{H\textsubscript{2} Price } - \text{Electrolyzer power} \times \text{Electrolyzer price} \times \Delta t
    \label{eq:eq3}
\end{equation}
   which directly encodes the net economic value of hydrogen production. Analysis confirms that the original controller never operated the electrolyzer  under grid deficit conditions (grid violation fraction = 0.0\%), indicating that physical feasibility is implicitly satisfied without requiring explicit penalty terms.

\subsubsection{Network Architecture and Training}

The actor network maps the 7-dimensional state to a normalized action via 7 → FC(256) → LayerNorm → ReLU → FC(128) → LayerNorm → ReLU → FC(1) → tanh. Layer normalization is preferred over batch normalization, as replay buffer mini-batches do not reflect the true marginal state distribution. The critic network maps the joint $(s,a)$ pair to a scalar $Q$-value via parallel state and action pathways, which are merged by element-wise addition and passed through a shared output head.\\
The offline training protocol preloads all 14{,}999 transitions into a replay buffer with capacity 20{,}000 prior to gradient updates. Exploration noise is suppressed during offline training to prevent deviation from the dataset policy. Key hyperparameters include: discount factor $\gamma = 0.99$, target network smoothing factor $\tau = 0.005$, mini-batch size $=256$, actor learning rate $= 1 \times 10^{-4}$, and critic learning rate $= 3 \times 10^{-4}$.
\section{Conclusion}

This study presented the first formally structured statistical and machine 
learning analysis of the Hydrogen Multi-Energy System (H-MES) dataset, 
applying a six-method analytical framework of exploratory data analysis, 
Kruskal--Wallis group comparison, multiple regression, and Random Forest 
variable importance in R Studio, complemented by Random Forest regression, 
LSTM sequence modeling and DDPG reinforcement learning in MATLAB to 
20,000 five-minute operational records spanning a full annual cycle.\\
The exploratory analysis revealed that H\textsubscript{2} production and 
electrolyzer operation is characterized by severe zero-inflation and bimodal 
distributions, reflecting the system's binary operating mode. This 
distributional structure, concealed by aggregate simulation reporting, has 
direct implications for statistical method selection and for understanding the 
true operational profile of P2H2P systems. The group comparison analysis 
produced the most operationally significant finding: solar PV availability 
acts as a threshold trigger rather than a continuous linear driver, with Low 
and Medium solar groups producing statistically indistinguishable H\textsubscript{2} 
output (Dunn's test: $z = 0.000$, $p = 1.000$) and meaningful electrolyzer  
activation occurring exclusively during peak solar periods 
($\varepsilon^2 = 0.457$, very large effect). Electricity demand exerted a 
weaker but significant monotonic inverse suppression effect 
($\varepsilon^2 = 0.126$), confirmed consistently across both R and MATLAB 
analyses. The regression model confirmed electrolyzer power input as the 
near-exclusive linear predictor ($\beta = 1.000$, $R^2 = 1.000$ on held-out 
test data), while the PV $\times$ Wind interaction was independently confirmed 
as significant ($F = 1{,}674.8$, $p < 0.001$). The strong 24-hour 
autocorrelation of electrolyzer  dispatch (lag-288 $r = 0.845$) motivated the 
LSTM's 24-hour look back window design, and the DDPG agent successfully 
translated these structural insights into a deployable continuous dispatch 
policy optimizing hydrogen revenue directly.\\
The principal novel contribution across both analytical streams is the 
demonstration that wind output carries unique non-linear predictive information 
(\%IncMSE rank 1\textsuperscript{st}, 45.16\%) undetectable by parametric 
regression, a finding absent from the simulation-based H-MES literature 
that provides a statistically grounded rationale for wind integration, even 
where linear correlations are weak ($r = 0.167$). The study's central 
methodological contribution is the demonstration that statistical and machine 
learning methods are genuinely complementary: statistical analysis reveals the 
causal and structural architecture of the H-MES system with inferential 
rigor, while machine learning operationalizes this understanding into 
predictive and optimization tools. Future work should extend this framework to 
real-world operational datasets across different system configurations and 
climatic conditions, incorporate time-series regression methods to formally 
address autocorrelation structure, and integrate the available economic 
variables toward integrated techno-economic modeling. 

\bibliographystyle{elsarticle-num}
\bibliography{Ref_Sa}

\end{document}